\documentclass{article} 
\usepackage{arxiv_style,times}

\usepackage{natbib}
\setcitestyle{aysep={,}}

\usepackage{amsmath,amsfonts,bm}

\def\eqref#1{equation~\ref{#1}}

\def\1{\bm{1}}

\DeclareMathAlphabet{\mathsfit}{\encodingdefault}{\sfdefault}{m}{sl}
\SetMathAlphabet{\mathsfit}{bold}{\encodingdefault}{\sfdefault}{bx}{n}

\usepackage{footmisc}

\definecolor{lightblue}{RGB}{54, 116, 230}
\usepackage[colorlinks,
            linkcolor=red,
            citecolor=lightblue,
            ]{hyperref}
\usepackage{url}
\usepackage{booktabs}
\usepackage{xcolor}         
\usepackage{booktabs}       
\usepackage{amsfonts}       
\usepackage{nicefrac}       
\usepackage{microtype}      
\usepackage{graphicx}
\usepackage{booktabs}
\usepackage{algorithm}  
\usepackage{array}   
\usepackage{colortbl}  
\usepackage{array}   
\usepackage{tikz}
\usepackage{comment}
\usepackage{amsmath,amssymb} 
\usepackage{multirow}
\usepackage{color}
\usepackage{bm} 
\usepackage{bbding}
\definecolor{myred}{RGB}{210, 10, 10}
\definecolor{mygreen}{RGB}{10, 210, 10}
\usepackage{authblk} 
\definecolor{lightgray}{rgb}{0.9, 0.9, 0.9}

\definecolor{baselinecolor}{gray}{.9}

\newlength\savewidth

\makeatletter
\def\blfootnote{\xdef\@thefnmark{}\@footnotetext}
\makeatother

\title{$\gamma-$MoD: Exploring Mixture-of-Depth Adaptation for Multimodal Large Language Models}

\author{
    Yaxin Luo$^{1}$,~Gen Luo$^{2}$$^\dagger$,~Jiayi Ji$^{3,4}$,~Yiyi Zhou$^{3}$,~Xiaoshuai Sun$^{3}$,~Zhiqiang Shen$^{5}$,~Rongrong Ji$^{3}$ \\ 
    ~$^1$Technical University Of Denmark \quad ~$^2$OpenGVLab, Shanghai AI Laboratory \quad \\$^3$Xiamen University \quad $^4$National University of Singapore \quad $^5$MBZUAI\\  
    \vspace{3mm}
    \textbf{Project Page:} \textcolor{magenta}\href{https://yaxin9luo.github.io/gamma-mod-webpage}{Gamma-MOD}
}
\arxivfinalcopy 
\begin{document}

\maketitle
\blfootnote{ $\dagger$Corresponding author.}
\begin{abstract}
Despite the significant progress in multimodal large language models (MLLMs), their high computational cost remains a barrier to real-world deployment. Inspired by the mixture of depths (MoDs) in natural language processing, we aim to address this limitation from the perspective of ``activated tokens''. Our key insight is that if most tokens are redundant for the layer computation, then can be skipped directly via the MoD layer. However, directly converting the dense layers of MLLMs to MoD layers leads to substantial performance degradation. To address this issue,  we propose an innovative MoD adaptation strategy for existing MLLMs called $\gamma$-MoD.  In $\gamma$-MoD,   a novel metric is proposed to guide the deployment of MoDs in the MLLM, namely \textit{rank of attention maps} (ARank). Through ARank, we can effectively identify which layer is redundant and should be replaced with the MoD layer.  Based on ARank,  we further propose two novel designs to maximize the computational sparsity of MLLM while maintaining its performance, namely \textit{shared vision-language router} and \textit{masked routing learning}.   With these designs, more than 90\% dense layers of the MLLM can be effectively converted to the MoD ones. To validate our method, we apply it to three popular MLLMs, and conduct extensive experiments on 9 benchmark datasets.  Experimental results not only validate the significant efficiency benefit of $\gamma$-MoD to existing MLLMs but also confirm its generalization ability on various MLLMs.  For example, with a minor performance drop, \emph{i.e.,} -1.5\%, $\gamma$-MoD can reduce the training and inference time of LLaVA-HR by 31.0\% and 53.2\%, respectively. 
\end{abstract}

\section{Introduction}
\label{sec:intro} 
Recent years have witnessed the great success of large language models (LLMs) in natural language processing (NLP)~\citep{GPT-4,LLaMA,internLM}, which  attracts increasing attentions in extending LLMs to vision-language (VL) tasks. Despite the progress, recent multimodal large language models (MLLMs)~\citep{LLaVA,llava-next,internvl1.5,flamingo} are often criticized by their expensive computational costs.  For example, the inference speed of existing MLLMs like LLaVA-HR~\citep{llava-hr} is still far from practical requirements,  \emph{e.g.,} 4.7 samples per second.  Driven by the progress of  NLP,   recent advances have employed the mixture-of-experts (MoEs)~\citep{moellava,mixtral}  to MLLMs to reduce the ``activated parameters'',  thus achieving trade-off between efficiency and performance. 

\begin{figure*}[t]
		\centering
		\includegraphics[width=1.\textwidth]{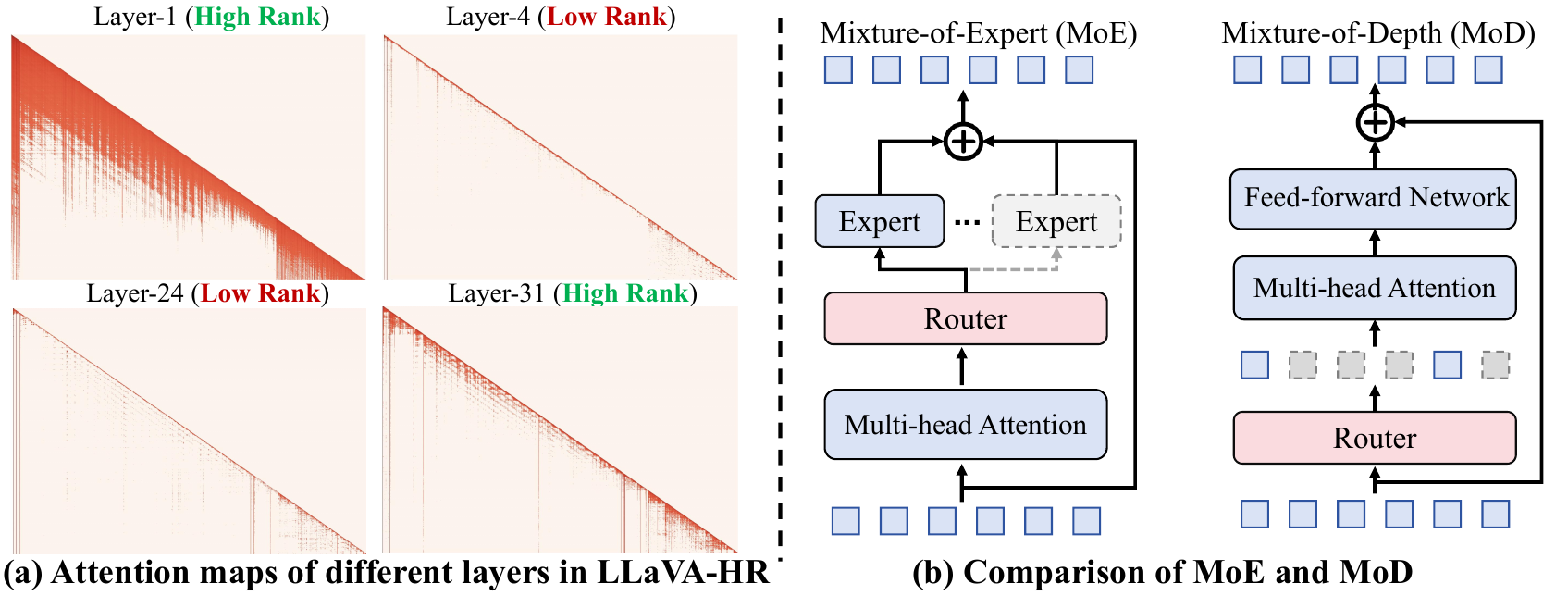} 
  		\vspace{-1.5em}
		\caption{\textbf{Visualization  of attention maps in the MLLM   and comparison of MoE with MoD.}    (a) Lower-rank layers often exhibit redundancy in their attention computation.  (b) 
  Different from MoE, MoD achieves the computational sparsity from the perspective of ``activated token'', where the computational budget is dynamically allocated to each token.
  }
		\label{fig1-1}
		\vspace{-1em}
	\end{figure*}

Orthogonal to MoEs, we aim to tackle the efficiency bottleneck of MLLMs from the perspective of ``activated tokens''.  As shown in Fig.~\ref{fig1-1} (a), a large number of  tokens are less important in the  computation, such as visual background and prepositional words. However, existing MoEs still allocate the same experts to all input tokens, leading to redundant computational costs.   A promising solution to this issue is the recently proposed mixture-of-depths (MoDs) in NLP~\citep{mod},  which equips each token with a router to determine whether a module should be computed.  However, recent MoDs~\citep{mod} typically require pre-training LLMs from scratch, and their employment on MLLMs still remains under-explored.

In this paper, we focus on the efficient adaptation of MoDs to existing MLLMs.  In particular, our goal is to maximize the computational sparsity of MLLMs while maintaining competitive performance.  However, directly converting all dense layers of MLLMs to MoD layers leads to significant performance degradation, \emph{e.g.,} -33.3\% of LLaVA-HR~\citep{llava-hr} on TextVQA~\citep{textvqa}.  In practice, we observe that such issue is mainly caused by two aspects.   Firstly, the deployment of MoDs lacks a practical guidance to measure the layer redundancy, thus undermining  the necessary dense layers.  As illustrated in Fig.~\ref{fig1-1} (a),  attention patterns vary significantly across layers, and some layers exhibit less redundancy.  Additionally, the setting of MLLMs,  \emph{e.g.,} input modality, differs substantially from that of LLMs, making the direct adaptation of MoDs suboptimal.

To overcome these limitations, we first propose a novel metric to guide the deployment of MoDs in MLLMs, called the \emph{rank of attention maps} (ARank).  Our key insight is that lower-rank attention maps indicate that fewer tokens are necessary for computation.  As shown in Fig.~\ref{fig1-1} (a),  most of tokens   of \textit{Layer-4} are assigned small attention weights, contributing minimally to the final output. This provides a valuable hint for us to replace the redundant layer  with the MoD one under the guidance of ARank.  In practice, the calculation of  ARank is both efficient and flexible.  Empirically, we  find that the average  ARank  always keeps the similar despite the change of  samples.  Therefore,   randomly sampling a small amount of data can already accurately estimate the ARanks.

Based on the ARank, we propose an innovative MoD adaptation strategy  for  existing MLLMs, called $\gamma$-MoD.  Specifically, $\gamma$-MoD is a plug-and-play approach that can be seamlessly integrated into current MLLMs via instruction tuning.  In  $\gamma$-MoD, two novel designs are adopted to maximize its   benefits to MLLMs, namely \textit{shared vision-language router} and \textit{masked routing learning}.  The shared vision-language router performs routing on the entire multimodal sequence and uses a weight-sharing strategy to facilitate optimization.  Then, masked routing learning is introduced to prevent critical tokens from being skipped during training, \emph{i.e.,} instruction tokens.  With these designs, over 90\% of dense layers can be converted to MoD layers with minimal performance sacrifice, resulting in even larger computational sparsity than the native MoD-based LLM~\citep{mod}.

To validate $\gamma$-MoD, we apply it to two popular MLLMs  and conduct extensive experiments on 9 vision-language benchmarks.  Experimental results show that  $\gamma$-MoD significantly improves the training and inference efficiency of existing MLLMs while keeping their  performance competitive.     For example, $\gamma$-MoD reduces 51.6\% Flops, 31\% training time and 53.2\% inference time for LLaVA-HR~\citep{llava-hr}, but its average performance decline is only -1.5\%.  More importantly, the great generalization ability of $\gamma$-MoD is also witnessed on different MLLM structures and parameter sizes. Overall, the contribution of the paper can be summarized in three folds:
\begin{itemize}
    \item We present a novel mixture-of-depth (MoD) framework for the sparse computation of existing MLLMs, namely $\gamma$-MoD, which can seamlessly convert most dense layers in MLLMs to the sparse MoD layers.
    \item We propose an innovative metric to measure the layer redundancy, namely rank of attention maps (ARank).  With ARank, we can best determine that which dense layer should be convert to the MoD one.
    \item We carefully explore the design of $\gamma$-MoD in existing MLLMs, including the shared vision-language router and the masked routing learning,  which can achieve up to 51.6\% computational sparsity with minor performance sacrifice. Extensive experiments also confirm the generalization ability of $\gamma$-MoD.
\end{itemize}
\section{Related Work}
\label{sec:related}
\subsection{Multimodal Large Language Models}

Large language models (LLMs)~\citep{GPT-4,LLaMA,mixtral,Falcon,internLM,phi3,SlimPajama-DC} have proven their strong capabilities in various natural language processing tasks~\citep{lambada,logicinfer,readingcomphrehension,helloswag}.
Motivated by this, numerous efforts~\citep{LLaVA,Qwen,mPLUG,InstructBLIP,internVL,miniGemini,Cambrian-1, llava-pp,dreamllm,showo,transfusion,minigpt,flamingo,emu2}  have been devoted into extending LLMs to multimodal large language models (MLLMs).   Among them, the most representative work is  LLaVA~\citep{LLaVA}, which uses a lightweight project to connect a visual encoder and an LLM. This simple framework has now become the de-facto paradigm in the community, empowering a set of MLLMs like Mono-InternVL~\citep{monointernvl}, Mini-Gemini~\citep{miniGemini} and InternVL~\citep{internVL}. Recently, researchers have shifted their attentions to high-resolution MLLMs. For example, LLaVA-NexT~\citep{llava-next} and InternVL-1.5~\citep{internvl1.5} adopt the dynamic image slicing strategy for high-resolution adaptation. LLaVA-HR~\citep{llava-hr} further propose a dual-branch structure to reduce the cost of high-resolution MLLMs. Despite the effectiveness, existing high-resolution MLLMs~\citep{llava-next,LLaVA-Onevision} will produce a much longer input tokens, resulting in prohibitively expensive computational costs. In this paper, the proposed $\gamma$-MoD can greatly overcome the efficiency bottleneck of existing MLLMs, which is significant for their practical applications.

\subsection{Sparse Computation for LLMs}
Existing LLMs has grown rapidly in their  parameter scale, which results in ever-increasing  computational costs~\citep{llama405B,K265B,Qwen72B,smaug72B,nemotron340B}.
Therefore, an influx of attentions have been focused on the sparse computation of LLMs.  Specifically, the mixture of experts (MoEs) are the most popular technology in the community~\citep{mm1,moesurvey,openmoe}, which dynamically activates part of expert networks for each token, thereby achieving trade-offs between capability and efficiency. For example,  Mixtra-8\texttimes7B~\citep{mixtral} and DeepSeekMoE~\citep{deepseekmoe}  replace the  feed-forward(FFN) layer of transformer block by an MoE Layer and the input tokens will be dynamically processed by top-K experts via the router.   
Orthogonal to MoE, \citet{mod} proposed the mixture of depths (MoDs) to dynamically allocate computations for each token. Compared to MoE,  the main principle of MoD is to reduce the ``activated tokens'' instead of the ``activated parameters''.  This paradigm has shown great potentials for the sparse computation of LLMs, but its potential   on MLLM is still under exploration. In the existing literature, most existing works aim to adapt MoEs to MLLMs.  For instance,  MoE-LLaVA~\citep{moellava} proposed a novel approach to convert a dense MLLM to a mixture-of-expert structure.  However, these methods often require additional training costs to realize the adaptation.  Orthogonal to these works, we are the first to explore MoDs on MLLMs, which can seamlessly realize  sparse computations for exiting MLLMs.


\section{Preliminaries}
We first recap the mechanism of \textit{Mixture of Experts} (MoEs) and \textit{Mixture of Depths} (MoDs). 

\textbf{Mixture of experts.} In particular, the main principle of MoE is to reduce the ``\textit{activated parameters}'' in dense models. Existing MoE-based LLMs~\citep{deepseekmoe,deepseekmoev2,moellava,mixtral} and MLLMs~\citep{llava-hr,internvl1.5,LLaVA} often contain multiple FFN modules in their layers, also termed \textit{experts}. During training and inference, only few experts are activated to participate in computations, thus retaining the trade-offs between performance and efficiency.  Given input features $x\in \mathbb{R}^{l\times d}$,  MoE  mechanism can be defined by
\begin{equation}
\begin{aligned} 
x&=x +\sum_{j=1}^k\mathcal{D}_{j}(x)R_j(x).
\label{eq_moe}
\end{aligned}
\end{equation}
Here, $\mathcal{D}(\cdot)$ denotes the expert layer, \emph{i.e.,} FFN. $k$ is the number of activated experts, and $R_j(\cdot)$ is the corresponding routing function.  In practice,  top-k experts are selected according to their routing scores, where $k$ is much smaller than the total number of experts $K$.  

\textbf{Mixture of depths.} Different from MoEs, MoDs aim to improve the model efficiency via the reduction of ``\textit{activated tokens}''.  Compared to MoEs, the routing mechanism of MoDs   performs on  input tokens, and most tokens will directly skip the  dense layer in MLLMs. Thus, MoDs can be written as
\begin{equation}
\begin{aligned} 
x_j = 
\begin{cases} 
x_j + \mathcal{D}(x_j)R(x_j) & \text{if } R(x_j)  \geq \delta_s, \\
x_j  & \text{if } R(x_j)  < \delta_s,
\end{cases}
\label{eq_mod}
\end{aligned}
\end{equation}
where $x_j \in \mathbb{R}^d$ denotes the token vector in $x$, and $\delta_s$ is a routing threshold. As defined in Eq.~\ref{eq_mod},  inactive tokens will directly skip the   layer $\mathcal{D}(\cdot)$ to save the computational cost.

\textbf{Discussion.} In  existing MLLMs~\citep{moellava}, MoE is typically used to efficiently scale up the model size, while its computations are not directly reduced. In contrast, MoD can perform as a plug-and-play module to save the cost of a common dense layer, which is more significant to the efficient scenario. Unfortunately, the adaptation of MoD to existing MLLMs is still under-explored, and its practical use in LLMs also requires expensive pretraining. 
 
\begin{figure}[t]
    \centering
    \includegraphics[width=1.\textwidth]{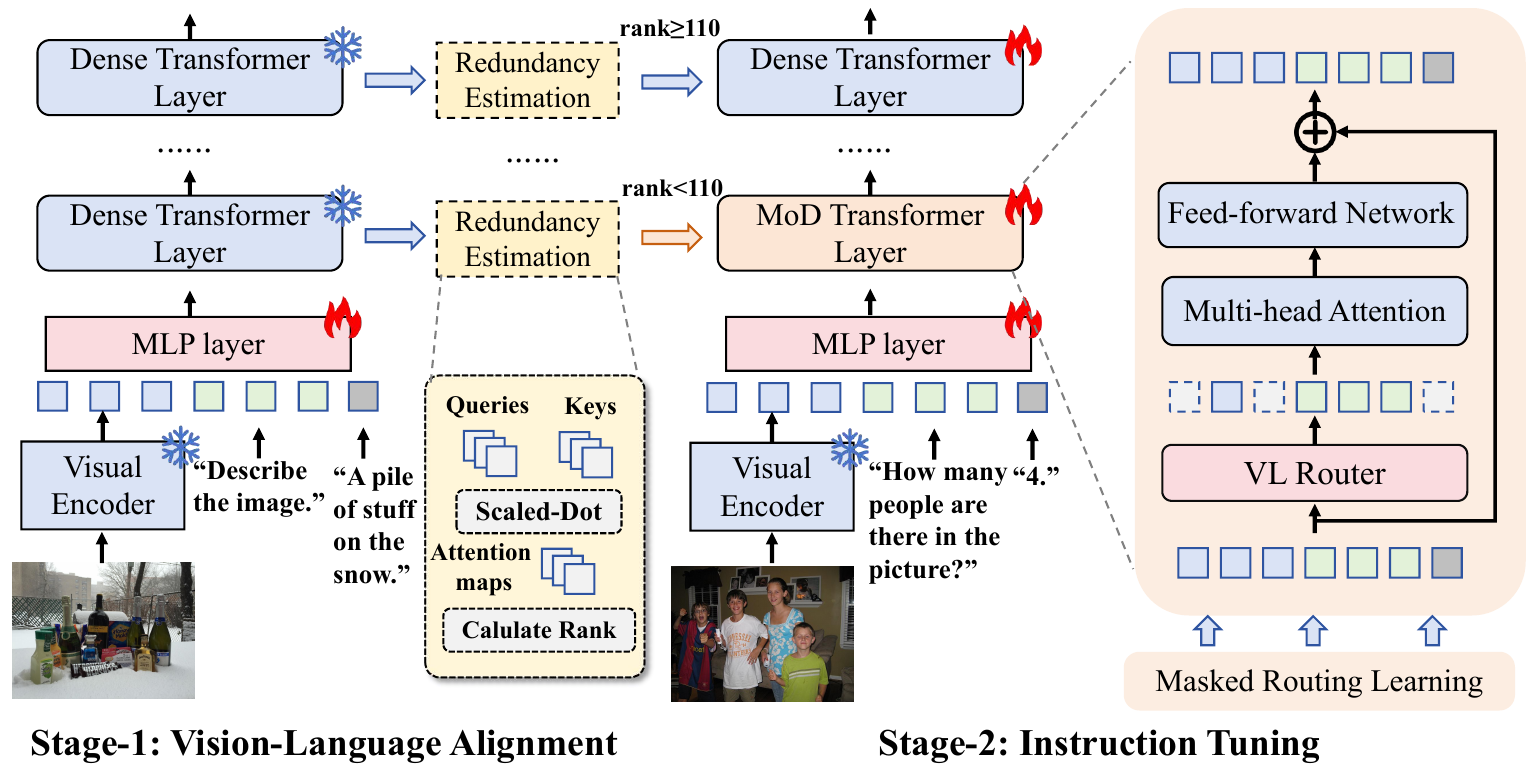}  
    \vspace{-2em}  
    \caption{\textbf{Illustration of  our $\gamma$-MoD adaptation on LLaVA-HR.} $\gamma$-MoD is a plug-and-play approach that can be directly applied in existing MLLMs. After vision-language alignment, $\gamma$-MoD can replace most redundant layers with MoD ones via the rank-based redundancy estimation. }
    \label{fig2}
    \vspace{-1em}  
\end{figure}

\section{Method}
\label{methods}
\subsection{Overview}
In this paper, we propose a novel method to efficiently deploy MoDs to existing MLLMs, namely $\gamma$-MoD.  
The core principle of $\gamma$-MoD is to identify  redundant MLLM layers via a novel metric called \textit{rank of attention maps} (ARank) and replace them with the proposed MoD layer. 
Therefore, the deployment of $\gamma$-MoD  in the   given MLLM, \emph{i.e., }$\mathcal{F}_{\text{MLLM}}(\cdot)$, can be formulated by
\begin{equation}
\begin{aligned} 
\mathcal{F}_{\text{MLLM}}&=\mathcal{G}_0\circ \mathcal{G}_1\circ \mathcal{G}_2...\circ \mathcal{G}_n, \\
\text{where} \quad \mathcal{G}_i&=
\begin{cases} 
\mathcal{D}_i \quad \text{if }  \tau(\mathcal{D}_i)  \geq \delta_{\tau}, \\
\mathcal{S}_i \quad  \text{if } \tau(\mathcal{D}_i)   < \delta_{\tau}.
\end{cases}
\label{eq_rmod}
\end{aligned}
\end{equation}
Here, $\mathcal{G}(\cdot)$ denotes the layer of the MLLM, where $\mathcal{S}(\cdot)$ and $\mathcal{D}(\cdot)$ indicate the  dense layer  and its MoD alternative, respectively.  $\tau(\cdot)$ is a function to estimate the  redundancy of the given dense layer $\mathcal{D}_i$, and $\delta_{\tau}$ is a threshold.   Given the architecture in Eq.~\ref{eq_rmod}, $\gamma$-MoD aims to maximize the sparsity while maintaining the performance. Thus, the optimization objective of $\gamma$-MoD can be written as:
\begin{equation}
\begin{aligned} 
&\arg \min_{\theta,\theta_r}\mathcal{L}_{obj}(\mathcal{F}_{\text{MLLM}}(x^0;\theta))+ \sum_{i=1}^k\mathcal{L}_{aug}(R(x^i;\theta_r)),\\
&s.t.  \quad \sum_{i=1}^{k}\sum_{j=1}^d \mathbb{I}_{R(x_j^i)  < \delta_s} = \alpha.
\label{eq_rmod_obj}
\end{aligned}
\end{equation}
Here, $\mathcal{L}_{obj}$ and $\mathcal{L}_{aug}$ denote the auto-regressive loss and the routing loss for the router $R(\cdot)$, respectively. $x^i$ is the input tokens  of \textit{i}-th  layer, and $\alpha$ is the pre-defined sparse target. $ \mathbb{I}_{R(x_j^i)  < \delta_s} \rightarrow \{0, 1\}$ is the indicator function, which is equal to 1 when $R(x_j^i)  < \delta_s$.

\subsection{Rank-based Redundancy Estimation} 
The key challenge of $\gamma$-MoD is how to identify the dense layer that should be converted to the MoD one.  In practice, directly replacing all layers with MoD ones will lead to significant performance degeneration. The original MoD-based LLM \citep{mod} overcomes this issue by the hand-craft attempt, which is still sub-optimal and time-consuming. 
However, in existing MLLMs, the LLM is already pre-trained on large scale of corpus, which can intuitively provide sufficient knowledge to achieve the process automatically.  

Motivated by this, we propose an innovative metric to estimate the token-wise redundancy of a layer in MLLM, namely \textit{rank of attention maps} (ARank). In particular, given   tokens $x^i \in \mathbb{R}^{l\times d}$ of $i$-th layer, ARank is defined by the average rank of attention maps:
\begin{equation}
\begin{aligned} 
&\tau(x^i,\mathcal{D}_i)= \frac{1}{n_h}\sum_{h=1}^{n_h}\text{rank}\big(A_h\big), \\
&\text{where} \quad A_h=(x^iW_Q^h)(x^iW_K^h)^T.
\label{eq_arank}
\end{aligned}
\end{equation}
 Here, $\text{rank}(\cdot)$ denotes the rank calculation.   $n_h$ is the number of attention heads.   $A_h \in \mathbb{R}^{l\times l}$ is the attention map in \textit{h}-th head, and $W_Q^h \in \mathbb{R}^{d\times\frac{d}{h}}$ and  $W_K^h \in \mathbb{R}^{d\times\frac{d}{h}}$ are the corresponding weights.

\begin{figure}[t]
    \centering
    \includegraphics[width=1.\textwidth]{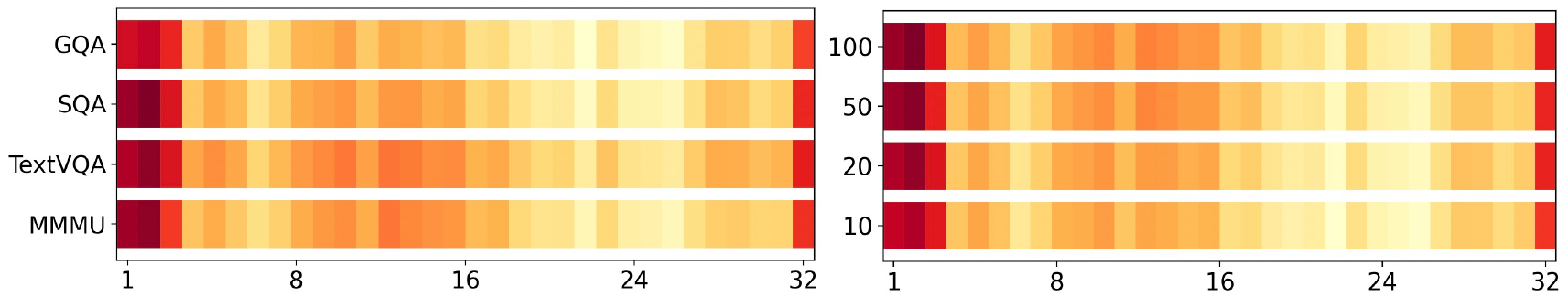}  
    \vspace{-2em}  
    \caption{\textbf{Visualization of ARank based on different tasks (left) and sample sizes (right).} The horizontal axis represents the layer index of LLaVA-HR. The darker color indicates the larger ARank.}
    \label{fig3}
    \vspace{-1em}  
\end{figure}
\textbf{Theoretical analysis of ARank.}  In Eq.~\ref{eq_arank}, attention map $A_h$ can well reflect the contribution of different tokens. Thus, $A_h$ with a low rank suggests that most tokens are less informative. To validate this, we conduct a SVD \citep{svd} analysis for $A_h$, which is written as
\begin{equation}
\begin{aligned} 
A_h&= \sum_{i=1}^r \sigma_i u_i v_i^T  =\sum_{i=1}^{r'} \sigma_i u_i v_i^T + \sum_{i=r'+1}^r \sigma_i u_i v_i^T,
\label{eq_arank_svd}
\end{aligned}
\end{equation}
where $r$ is the rank of $A_h$ and  $r' \ll r$ is a constant value. $\sigma_i$, $u_i$ and  $v_i$ denote the \textit{i}-th single value, left single vector and right single vector of $A_h$, respectively.  As shown in Eq.~\ref{eq_arank_svd}, $A_h$ can be deposed to a matrix of rank $r'$ and additional information, \emph{i.e.,} $\sum_{i=r'+1}^r \sigma_i u_i v_i^T$.  Therefore, lower-rank attention map  suggests   higher redundancy, which implies that MoD can be deployed to skip most tokens.

  \noindent \textbf{Practical calculation of ARank.}  As defined in Eq.~\ref{eq_arank}, the calculation of ARank is highly dependent on input samples.  Therefore, it is still challenging to accurately calculate the ARank due to the variance of individual samples.  Inspired by HRank~\citep{HRank}, we estimate ARank using its expectation on a batch of samples, which is practically robust.  As shown in Fig.~\ref{fig3}, we visualize  average  ARank values of LLaVA-HR~\citep{llava-hr} based on different samples.  From these results, we empirically find that the expected ARank remains largely consistent across different tasks. Therefore, a small batch of samples is sufficient to accurately calculate ARank. In our experiments, we set the sample size to 50, ensuring computational efficiency.

\subsection{Mixture-of-Depth Adaptation}  
To maximize the effectiveness of MoDs to existing MLLMs, we carefully investigate the micro design of MoDs, including the shared vision-language router and the  masked routing learning.

\noindent \textbf{ Shared vision-language router.} Conventional MoDs~\citep{mod} are designed for LLMs, so their routing is only performed on textual tokens. In MLLMs, such a strategy is sub-optimal due to the large redundancy of visual tokens~\citep{efficient_mllm,tokenfusion,internvl1.5,LLaVA-Onevision}. Therefore, the router of $\gamma$-MoD, \emph{i.e.,} $R(\cdot)$, aims to skip both visual and textual tokens, which is defined by
\begin{equation}
\begin{aligned} 
R(x)=\text{softmax}(xW_R+b_R),
\label{eq_router}
\end{aligned}
\end{equation}
where $x=\{q,a,t\}$ denotes the vision-language tokens, which consist of question tokens $q \in \mathbb{R}^{l_q\times d }$, image tokens $a \in \mathbb{R}^{l_a\times d }$ and textual response tokens $t \in \mathbb{R}^{l_t\times d }$.  $W_R \in \mathbb{R}^{l\times 2}$ and $b_R\in \mathbb{R}^{2}$ are the weights and bias, respectively. Notably, we use a binary softmax function to produce the routing probability, where $R(x)^0$ denotes the probability of skipping. Based on Eq.~\ref{eq_router}, we further share the router parameters for all MoD layers, which is significant for the stable optimization. To explain,  the shared router  receives more gradients from different layers, greatly facilitating its convergence at the beginning of training. 

\noindent \textbf{Masked routing learning.} During VL training, not all tokens contribute equally to the optimizing process.  In particular, the skip of key tokens in the question, \emph{e.g.,} subject, will greatly hurt the generative training as the answer relies on these conditional elements. Therefore, we introduce a masked routing learning strategy to prevent these tokens from being dropped during training. In this case, the objective of the routing learning can be defined by 
\begin{equation}
\begin{aligned} 
\mathcal{L}_{aug}(x)=\big(R(x)^1\cdot M_q \big) \log (\hat{R})+  \big(1- R(x)^0\cdot M_q \big) \log (1- \hat{R}). 
\label{eq_router_loss}
\end{aligned}
\end{equation}
Here, $M_q \in \mathbb{R}^{l\times1}$ denotes the binary mask, where the question tokens are assigned to 0. $\hat{R} \in $ is the one-hot vector, where the position with top-k routing scores are assigned to 1. 

\noindent \textbf{The training scheme. }   $\gamma$-MoD is a plug-and-play approach for existing MLLMs, and the training scheme of existing MLLMs does not necessarily need to be carefully adjusted. In particular, the training of existing MLLMs can be roughly divided into two stages: vision-language alignment and instruction tuning.  After VL alignment, $\gamma$-MoD estimates the  redundancy of  a layer  using ARank, and directly replaces the redundant one with our MoD layer.  During instruction tuning, the routing parameters are jointly optimized with the routing and task objectives via Eq.~\ref{eq_rmod_obj}.  Notably,  other training configurations can simply remain the same as the original setting of MLLM.

\section{Experiments}
In this section, we provide extensive ablation studies to analyze the key designs that contribute to the effectiveness of our proposed $\gamma$-MoD framework. We also evaluate $\gamma$-MoD on multiple benchmarks and variant settings with existing dense and sparse MLLMs.
\label{experiments}
\subsection{Datasets and Metrics} 
 We evaluate our $\gamma$-MoD on five multimodal benchmarks for MLLMs, which includes POPE~\citep{pope}, MME~\citep{MME}, MMB~\citep{MMB}, MMMU~\citep{MMMU} and MM-Vet~\citep{mmvet}. Specifically, POPE and MM-Vet aim to evaluate the visual hallucinations of MLLMs. MME measures both perception and cognition abilities on a total of 14 subtasks \emph{e.g.,} numerical calculation, text translation, and code reasoning. MMBench is a structured objective benchmark designed for comprehensive and robust evaluation of vision-language models. MMMU is designed to measure the perception, knowledge, and reasoning of MLLMs' abilities. We report all the results in their default settings. For MME, we  report the perception score.

\begin{table}[!t]
\centering
\caption{
\textbf{Comparison of different $\gamma$-MoD configurations  on LLaVA-HR.} The default setting used in the table is colored in gray. ``Q'' and ``A'' refer to question and answer tokens.
}
\setlength\tabcolsep{3pt}
\resizebox{\linewidth}{!}{
\begin{tabular}{l|cc|cc|cc|cc|ccc}
\toprule[1.2pt]
\multicolumn{1}{c|}{\multirow{2}{*}{Methods}} & \multicolumn{2}{c|}{GQA} & \multicolumn{2}{c|}{SQA} & \multicolumn{2}{c|}{MMMU} & \multicolumn{2}{c|}{TextVQA}  & \multicolumn{3}{c}{Average} \\
\multicolumn{1}{c|}{} & \multicolumn{1}{c}{Acc.}  & \multicolumn{1}{c|}{Skip} & \multicolumn{1}{c}{Acc.} & \multicolumn{1}{c|}{Skip} & \multicolumn{1}{c}{Acc.} & \multicolumn{1}{c|}{Skip} & \multicolumn{1}{c}{Acc.}& \multicolumn{1}{c|}{Skip}  & \multicolumn{1}{c}{Acc.} & \multicolumn{1}{c}{TFlops} & \multicolumn{1}{c}{Skip}\\ 
\midrule[0.4pt]
LLaVA-HR~\citep{llava-hr}       &64.2 & 0\% & 67.9 & 0\% & 34.6 & 0\% & 67.1 & 0\% & 58.5  & 19.2 & 0\% \\
\midrule[0.4pt]
\textit{MoD layer:} & \multicolumn{1}{l}{} & \multicolumn{1}{l|}{} & \multicolumn{1}{l}{} & \multicolumn{1}{l|}{} & \multicolumn{1}{l}{} & \multicolumn{1}{l|}{} & \multicolumn{1}{l}{}& \multicolumn{1}{l|}{}& \multicolumn{1}{l}{} & \multicolumn{1}{l}{} & \multicolumn{1}{l}{}\\ 
All layers  & 45.9 & 38.2\%  & 42.6 & 33.7\% & 25.9 & 32.8\% & 33.8 & 34.1\% & 37.1 & 12.3 & 34.7\% \\
1 MoD per 2 layers & 57.8 & 19.1\% & 52.3 & 16.5\% & 26.9 & 16.6\% & 54.0 & 17.9\% & 47.8 & 16.1 & 17.5\% \\
2 MoDs per 3 layers & 38.1 & 26.8\% & 46.5 & 24.6\% & 24.3 & 24.4\% & 42.1 & 24.9\% & 37.8 & 15.9 & 25.2\% \\
\rowcolor{baselinecolor}
ARank-based deployment& 63.7 & 40.7\%  & 68.5 & 35.9\% & 35.6 & 36.8\% & 65.3 & 38.2\% & 58.3 & 12.6 & 37.9\%\\
 \midrule[0.4pt]
\textit{Masked token:} & \multicolumn{1}{l}{} & \multicolumn{1}{l|}{} & \multicolumn{1}{l}{} & \multicolumn{1}{l|}{} & \multicolumn{1}{l}{} & \multicolumn{1}{l|}{} & \multicolumn{1}{l}{}& \multicolumn{1}{l|}{}& \multicolumn{1}{l}{} & \multicolumn{1}{l}{} & \multicolumn{1}{l}{}\\
None & 63.2 & 52.0\% & 66.8 & 46.9\% & 33.9 & 47.0\% & 64.7 & 49.8\% & 57.2 & 10.7 & 48.9\%\\
\rowcolor{baselinecolor}
Q & 63.7 & 40.7\%  & 68.5 & 35.9\% & 35.6 & 36.8\% & 65.3 & 38.2\% & 58.3 & 12.6 & 37.9\% \\
Q + A & 62.8 & 38.8\%  & 68.6 & 30.5\% & 34.7 & 35.4\% & 62.0 & 37.2\% & 57.0 & 13.0 & 35.5\% \\
\midrule[0.4pt]
\textit{Shared router:} & \multicolumn{1}{l}{} & \multicolumn{1}{l|}{} & \multicolumn{1}{l}{} & \multicolumn{1}{l|}{} & \multicolumn{1}{l}{} & \multicolumn{1}{l|}{} & \multicolumn{1}{l}{}& \multicolumn{1}{l|}{}& \multicolumn{1}{l}{} & \multicolumn{1}{l}{} & \multicolumn{1}{l}{}\\
Not Share & 60.6 & 55.8\% & 64.5 & 48.2\% & 32.1 & 48.9\% & 58.4 & 52.9\% & 53.9 & 10.3 & 51.5\%\\
\rowcolor{baselinecolor}
Share & 63.1 & 60.3\%  & 67.9 & 56.9\% & 34.7 & 56.6\% & 64.9 & 57.1\% & 57.6 & 9.3 & 57.7\% \\
\midrule[0.4pt]
\textit{Routing ratio:} & \multicolumn{1}{l}{} & \multicolumn{1}{l|}{} & \multicolumn{1}{l}{} & \multicolumn{1}{l|}{} & \multicolumn{1}{l}{} & \multicolumn{1}{l|}{} & \multicolumn{1}{l}{}& \multicolumn{1}{l|}{}& \multicolumn{1}{l}{} & \multicolumn{1}{l}{} & \multicolumn{1}{l}{}\\
17\%  & 63.6 & 18.9\%  & 68.9 & 15.5\% & 34.7 & 14.7\% & 66.1 & 16.5\% & 58.3 & 16.3 & 16.4\% \\
\rowcolor{baselinecolor}
34\%  & 63.7 & 40.7\%  & 68.5 & 35.9\% & 35.6 & 36.8\% & 65.3 & 38.2\% & 58.3 & 12.6 & 37.9\% \\
51\% & 63.1 & 60.3\%  & 67.9 & 56.9\% & 34.7 & 56.6\% & 64.9 & 57.1\% & 57.6 & 9.3 & 57.7\% \\
68\%  & 59.1 & 77.8\% & 70.1 & 73.5\% & 33.7 & 71.8\% & 58.4 & 74.1\% & 55.3 & 6.5 & 74.3\% \\
\bottomrule[1.2pt]
\end{tabular}%
}
\label{mod_config} 
  \vspace{-1em}
\end{table}
We also evaluate $\gamma$-MoD on six image question answering benchmarks,VQAv2~\citep{vqav2}, VizWiz~\citep{vizwiz}, TextVQA~\citep{textvqa}, SQA~\citep{sqa}, GQA~\citep{gqa} and SEED~\citep{seed}. In particular, SQA~\citep{sqa} and VizWiz~\citep{vizwiz} are two zero-shot tasks, and none of their samples are present in our training data. We report the overall accuracy of SEED, the test set of VizWiz and we organize samples of these tasks in instruction formats of LLaVA-1.5~\citep{llava1.5}.
\subsection{Implementation Details}
For all models, pre-training is conducted on  LCS-558K dataset~\citep{llava1.5}, which includes high-quality 558k image-text pairs. For instruction tuning, we follow  LLaVA-1.5~\citep{llava1.5} to use 665k vision-language instruction data. 
To deploy $\gamma$-MoD to MLLMs,  ARank is  calculated to identify redundant layers  after the pre-training stage. For all models,  the fourth largest ARank value is used as the threshold for converting dense layers to MoD ones.   During instruction tuning, the coefficient for the routing loss is set to 0.01.  The remaining settings are kept the same with LLaVA-HR~\citep{llava-hr} and LLaVA~\citep{llava1.5}, including learning rate, training epochs, optimizer and datasets, \textit{etc}.
\subsection{Experimental Results}

\subsubsection{Quantitative analysis}

\noindent \textbf{Comparison with different MoD configurations.} In Tab.~\ref{mod_config}, we first compare different settings of MoD on LLaVA-HR~\citep{llava-hr}. From this table,  the first observation is that directly converting all layers to MoD ones leads to worse results, \emph{e.g.,} 33.8\% on TextVQA. Besides, although the hand-craft strategy performs much better,  its performance declines are still obvious, \emph{e.g.,} -10.7\% of  1 MoD per 2 layers on average.    These results confirm the challenges of adopting MoDs to MLLMs.   However, after employing our ARank-based strategy, the efficiency of LLaVA-HR is greatly increased while the performance is well maintained.   Such comparison greatly confirm the effectiveness of our ARank-based strategy against these baselines.

\begin{table}[t]
\caption{
\textbf{Ablation study of $\gamma$-MoD on LLaVA-HR.} 
``\emph{Param}'', ``\emph{Acc.}'' and  ``\emph{Skip}''  indicate the parameter, accuracy, and skip ratio, respectively. 
}
\vspace{2mm}
\centering
\setlength{\tabcolsep}{3pt}
\resizebox{1.0\textwidth}{!}
{
\begin{tabular}{lc|cc|cc|cc|cc|ccc}
\toprule[1.00pt]
\multicolumn{1}{l}{\multirow{2}{*}{Methods}} &\multicolumn{1}{c|}{\multirow{2}{*}{Param}} & \multicolumn{2}{c|}{GQA} & \multicolumn{2}{c|}{SQA} & \multicolumn{2}{c|}{MMMU} & \multicolumn{2}{c|}{TextVQA}  & \multicolumn{3}{c}{Average}  \\
\multicolumn{1}{c}{} & \multicolumn{1}{c|}{}& \multicolumn{1}{c}{Acc.} & \multicolumn{1}{c|}{Skip} & \multicolumn{1}{c}{Acc.} & \multicolumn{1}{c|}{Skip} & \multicolumn{1}{c}{Acc.} & \multicolumn{1}{c|}{Skip} & \multicolumn{1}{c}{Acc.} & \multicolumn{1}{c|}{Skip}&\multicolumn{1}{c}{Acc.} & \multicolumn{1}{c}{TFlops}  & \multicolumn{1}{c}{Skip}  \\ 
\midrule
LLaVA-HR~\citep{llava-hr}     &7.4B &64.2 & 0\% & 67.9 & 0\% & 34.6 & 0\% & 67.1 & 0\% & 58.5  & 19.2 & 0\% \\
\midrule
+ Default MoD~\citep{mod}  &7.4B & 45.9 & 38.2\%  & 42.6 & 33.7\% & 25.9 & 32.8\% & 33.8 & 34.1\% & 37.1 & 12.3 & 34.7\% \\
+ ARank-based deployment (ours) &7.4B& 63.2 & 52.0\% & 66.8 & 46.9\% & 33.9 & 47.0\% & 64.7 & 49.8\% & 57.2 & 10.7 & 48.9\%\\
\rowcolor{baselinecolor}
+ Masked routing learning (ours) &7.4B& 63.1 & 60.3\%  & 67.9 & 56.9\% & 34.7 & 56.6\% & 64.9 & 57.1\% & 57.6 & 9.3 & 57.7\%\\
\bottomrule[1.00pt]
\end{tabular}
\label{components}
}

\end{table}

In Tab.~\ref{mod_config}, we also validate  different micro-designs for deploying MoD on MLLM, including the masked routing learning, the shared router and the routing ratio.  From these comparisons, we first see that the masked learning strategy is much beneficial to  the optimization of $\gamma$-MoD, providing up to +1.7\% gains on SQA.  Without this strategy,   question tokens will be dropped in MoD layers, easily resulting in the semantic ambiguity for answering.  In addition, we also find that the router sharing strategy plays a significant role in $\gamma$-MoD. After removing this strategy, model performance will obviously drop on TextVQA by -6.5\%.  Finally, we validate the impact of different routing ratio on LLaVA-HR. From results we can see that model performance can be retained under  relatively small routing ratios, \emph{i.e.,} 17\% and 34\%. When  routing ratio is increased to $51\%$, model performance drops slightly from 58.3\% to 57.6\% on average. However, the benefit of efficiency is still notable, \emph{i.e.,} -51.5\% Flops. Overall, these comparisons greatly validate our motivations and the design of $\gamma$-MoD.

\begin{table}[!t]
\centering
\caption{
\textbf{ Results of $\gamma$-MoD on different MLLM architectures and model scales. }  $\gamma$-MoD-0.3 and $\gamma$-MoD-0.5 denote the routing ratio of 30\% and 50\%, respectively.
}
\vspace{.5em}
\setlength\tabcolsep{5pt}
\resizebox{\linewidth}{!}{
\begin{tabular}{lc|cc|cc|cc|cc|ccc}
\toprule[1.2pt]
\multicolumn{1}{c}{\multirow{2}{*}{Methods}} &\multicolumn{1}{c|}{\multirow{2}{*}{Param}} & \multicolumn{2}{c|}{GQA} & \multicolumn{2}{c|}{SQA} & \multicolumn{2}{c|}{MMMU} & \multicolumn{2}{c|}{TextVQA}  & \multicolumn{3}{c}{Average} \\
\multicolumn{1}{c}{} & \multicolumn{1}{c|}{} & \multicolumn{1}{c}{Acc.}  & \multicolumn{1}{c|}{Skip} & \multicolumn{1}{c}{Acc.} & \multicolumn{1}{c|}{Skip} & \multicolumn{1}{c}{Acc.} & \multicolumn{1}{c|}{Skip} & \multicolumn{1}{c}{Acc.}& \multicolumn{1}{c|}{Skip}  & \multicolumn{1}{c}{Acc.} & \multicolumn{1}{c}{TFlops} & \multicolumn{1}{c}{Skip}\\ 
\midrule[0.4pt]\midrule[0.4pt]
 \multicolumn{2}{l}{\textit{MLLM architecture:} }  \\ 
LLaVA &7B& 62.0 & 0\% & 66.8  & 0\% & 34.3 & 0\% & 58.2 & 0\%   & 55.3 & 10.7 & 0\% \\ 
 +$\gamma$-MoD-0.3  &7B& 61.1 & 34.1\% & 64.7 & 29.4\% & 35.4 & 29.8\% & 56.3 & 30.7\%& 54.4 & 7.7 & 31.0\%  \\
  +$\gamma$-MoD-0.5   &7B& 41.4 & 60.9\% & 62.3  & 54.8\% & 31.0 & 53.6\% & 42.9 & 56.2\%   & 44.4 & 5.3 & 56.4\% \\ \midrule
LLaVA-HR&7B &64.2 & 0\% & 67.9 & 0\% & 34.6 & 0\% & 67.1 & 0\% & 58.5  &19.2 & 0\%\\

+$\gamma$-MoD-0.3 &7B& 63.7 & 40.7\%  & 68.5 & 35.9\% & 35.6 & 36.8\% & 65.3 & 38.2\% & 58.3 & 12.6 & 37.9\%       \\
 +$\gamma$-MoD-0.5 &7B& 63.1 & 60.3\%  & 67.9 & 56.9\% & 34.7 & 56.6\% & 64.9 & 57.1\% & 57.6 & 9.3 & 57.7\%\\ 
 \midrule[0.4pt]
Mini-Gemini-HD &7B & 62.9 & 0\% & 69.6  & 0\% & 36.8 & 0\% & 66.5 & 0\% & 59.0 & 60.2 & 0\%  \\

+$\gamma$-MoD-0.3&7B& 62.1 & 37.1\% & 69.0  & 34.6\% & 34.1 & 36.4\% & 66.4 & 36.6\%  & 57.9 & 39.4 & 36.2\%  \\
+$\gamma$-MoD-0.5 &7B & 62.2 & 59.2\% & 70.4  & 56.8\% & 33.9 & 58.6\% & 67.0 & 57.7\%  & 58.4 & 27.8 & 58.1\% \\
\midrule[0.4pt]
\textit{Model scales:}  \\ 

LLaVA-HR &7B &64.2 & 0\% & 67.9 & 0\% & 34.6 & 0\% & 67.1 & 0\% & 58.5  &19.2 & 0\%\\ 

+$\gamma$-MoD-0.3 &7B& 63.7 & 40.7\%  & 68.5 & 35.9\% & 35.6 & 36.8\% & 65.3 & 38.2\% & 58.3 & 12.6 & 37.9\%      \\
  +$\gamma$-MoD-0.5 &7B& 63.1 & 60.3\%  & 67.9 & 56.9\% & 34.7 & 56.6\% & 64.9 & 57.1\% & 57.6 & 9.3 & 57.7\% \\ \midrule
LLaVA-HR & 13B & 64.8 & 0\%  & 68.1 & 0\% & 36.7 &0\%  & 68.1 & 0\% & 59.4 & 37.1  & 0\% \\ 

+$\gamma$-MoD-0.3  & 13B & 64.5 & 38.1\%  & 70.5 & 33.1\% & 37.8 & 32.5\% & 67.0 & 36.0\% & 60.0 & 25.1 & 34.9\%  \\
 +$\gamma$-MoD-0.5 &13B& 64.8 & 58.8\% & 69.5 & 52.2\% & 35.8 & 53.8\% & 66.8 & 55.4\% & 59.2 & 18.4 & 55.1\% \\ 
\bottomrule[1.2pt]
\end{tabular}%
}
\label{genralization} 
  \vspace{-1em}
\end{table} 

\noindent \textbf{Ablation studies.} To validate  contributions of each design in $\gamma$-MoD, we conduct ablation study in Tab.~\ref{components}. From this table, we can see that the default MoD will cause obvious performance degeneration, resulting up to -25.3\% on SQA.  In stark contrast, with our ARank-based deployment, the average performance of LLaVA-HR  is improved from 37.1\% to 57.6\%, and the computational sparsity also boosts from 34.7\% to 48.9\%.  Such comparison confirms that not all layers can be converted to MoD layers, and ARank is critical to identify the redundant ones. In addition, the use of masked routing learning can further benefit the model training, providing +0.8\% on MMMU and +0.2\% on TextVQA, respectively.  It worth noticing that the masked routing learning also increases the efficiency of $\gamma$-MoD, where the average computational costs are further reduced from 10.7 TFlops to 9.3 TFlops.  These results further confirm the effectiveness   of $\gamma$-MoD.

\noindent \textbf{Generalizations of $\gamma$-MoD on different MLLMs.} 
 In Tab.~\ref{genralization}, we also evaluate the generalization capability of $\gamma$-MoD across different MLLM architectures and model scales.  
 In particular, $\gamma$-MoD with 30\% routing ratio  demonstrates great trade-off between performance and efficiency on LLaVA. When the routing ratio increases to 51\%, the performance of LLaVA decreases significantly, suggesting its relatively low tolerance to high routing ratio.
 For LLaVA-HR, the $\gamma$-MoD-0.3 configuration maintains high accuracy  63.7\% on GQA and 65.3\% on TextVQA  while reducing TFlops by 34\% and skipping 37.9\% of tokens. When the routing ratio increases to 51\%, the token skip rate improves to 57.7\%, though a slight drop in accuracy is observed \emph{e.g.,} -0.6\% on GQA. These comparisons also reflect that high-resolution MLLMs often have a higher token redundancy than low-resolution ones.
When scaling to larger models, such as the LLaVA-HR-13B, our method continues to perform strongly. The $\gamma$-MoD-0.3 configuration yields a 38.1\% skip rate and 25.1 TFlops with minimal accuracy loss, suggesting that larger models are better suited to handle higher skip rates while maintaining performance. Even increasing  the routing ratio to 51\%  the competitive accuracy is still maintained,  \emph{e.g.,} 64.8\% on GQA and 66.8\% on TextVQA.

\begin{table*}[t]
\caption{\textbf{Training and inference efficiency of $\gamma$-MoD on LLaVA-HR.} The inference efficiency is tested on an NVIDIA A100 GPU, which is the average value of GQA, SQA, MMMU, and TextVQA.  }
\centering
\renewcommand\arraystretch{1.25}
\setlength\tabcolsep{16pt}
\resizebox{\linewidth}{!}{
\begin{tabular}{lccccccc}
\toprule
\multicolumn{1}{c}{\multirow{2}{*}{Methods}} & Training & Inference & Inference& Inference &Avg. \\  
\multicolumn{1}{c}{}                         & Time $\downarrow$              &  Throughput $\uparrow$        & Memory $\downarrow$       & TFlops $\downarrow$        & Acc. $\uparrow$      \\ \hline
LLaVA-HR                                     &  20.7 h                &       4.7 samples/s       &        19 G      &        19.2        &      58.5       \\
+$\gamma$-MoD-0.3                            &    15.4 h              &         5.9 samples/s        &       15 G       &         12.6        &      58.3       \\
+$\gamma$-MoD-0.5                            &     14.3 h             &        7.2 samples/s         &       14 G       &      9.3         &      57.6       \\ \hline
Gains                                    & \textcolor[rgb]{0,0.6,0}{\textbf{-31.0\% } }           &        \textcolor[rgb]{0,0.6,0}{\textbf{ +53.2\% }}       &          \textcolor[rgb]{0,0.6,0}{\textbf{ -26.3\%}}   &        \textcolor[rgb]{0,0.6,0}{\textbf{ -51.6\% }}      &    \textcolor[rgb]{0.6,0,0}{\textbf{  -1.5\%}}       \\ \bottomrule
\end{tabular}}
\label{efficiency}
\end{table*}

\begin{table*}[!t]
\small
\setlength\tabcolsep{0.85mm}
\caption{\textbf{Comparison with existing dense and sparse MLLMs on 9 benchmarks.} Speed is   the average samples per second of GQA, SQA, MMMU, and TextVQA. }
\vskip 0.1in
\label{tab:main_res}
\centering
\resizebox{\linewidth}{!}{
\begin{tabular}{l|c|cccc|ccccc|c}
\toprule
\multirow{2}{*}{\textbf{Methods}} & \multirow{2}{*}{\textbf{Param.}} & \multicolumn{4}{c|}{\textbf{Image Question Answering}} & \multicolumn{5}{c|}{\textbf{Benchmark Toolkit}}& \multirow{2}{*}{\textbf{Speed}} \\
& & TextVQA& VQA$^\text{v2}$ & GQA & SQA$^\text{I}$  & POPE & MME & MMB & MMMU & MM-Vet &   \\
\midrule
\multicolumn{10}{l}{\textit{Dense Model:}} \\
I-80B ~\citep{i80B}  & 65B & - &60.0 &45.2 &  -  &  - & - &54.5 & - & - & - \\
InstructBLIP~\citep{InstructBLIP} & 14B  &50.7 & - & 49.5 & 63.1 & 78.9 & 1212.8 & - & - & 25.6 & -\\
VILA~\citep{lin2024vila} & 7B & 64.4 & 79.9 & 62.3 & 68.2 & 85.5 & 1533.0 & 68.9 & - & 34.9 & -\\
{Qwen-VL~\citep{qwenvl} } &  10B & 63.8 &  78.8 &  59.3  &  67.1 & - &  1487.6 &  38.2 &  - &  - & 4.6 \\
 {LLaVA-1.5~\citep{llava1.5} }  &   7B  & 58.2 &   78.5 &  62.0 & 66.8 & 85.9 &   1510.7 &   64.3 &  34.3 &   30.5 & 8.1 \\ 
  LLaVA-HR~\citep{llava-hr}  &    7B  &67.1  & 81.9 & 64.2 &  67.9 & 87.6 & 1554.9 &  66.8  &  35.2 &  31.2  &  4.7 \\
   LLaVA-HR~\citep{llava-hr}  &    13B   & 68.1 & 82.3 &  64.8 & 68.1  & 87.8 &  1540.9  &  64.5  &  36.3  &  34.8 & 3.1\\
\midrule

\multicolumn{11}{l}{\textit{Sparse Model:}} \\
MoE-LLaVA~\citep{moellava}  & 3B & 50.1 & 76.7 & 60.3 & 62.6 & 85.7 & 1318.2 & 60.2 & - & 26.9 & 8.5 \\
MoE-LLaVA~\citep{moellava} & 5B & 51.4 & 77.6 & 61.4 & 68.5 & 86.3 & 1423.0 & 65.2 & - & 34.3 & 5.6 \\
\rowcolor{baselinecolor} $\gamma$-MoD-LLaVA(ours) &7B & 56.3 & 77.6 & 61.1 & 64.7 & 86.0 & 1342.1 & 59.4 & 35.4 & 29.8 & 10.3 \\
\rowcolor{baselinecolor} $\gamma$-MoD-LLaVA-HR(ours) &7B & 64.9 & 80.6 & 63.1 & 67.9 & 87.3 & 1516.0 & 63.4 & 34.7 & 31.5 & 7.2  \\
\rowcolor{baselinecolor} $\gamma$-MoD-LLaVA-HR(ours) &13B & 66.8 & 82.0 & 64.8 & 69.5 & 86.7 & 1515.4 & 65.2& 35.8 & 34.0 & 4.8\\
\bottomrule
\end{tabular}
\label{compare-sota}
}
\vskip -0.1in
\end{table*}

\noindent \textbf{Efficiency analysis.} In Tab.~\ref{efficiency}, we compare the training and inference efficiency of $\gamma$-MoD on LLaVA-HR. From  these results, we observe  comprehensive advantages of $\gamma$-MoD in terms of training and inference inference.   In particular,  $\gamma$-MoD-0.3 already achieves an obvious improvement in efficiency, \emph{i.e.,} -26\% training time and -35\% TFlops. However, the performance drops of $\gamma$-MoD-0.3  can be almost ignorable, \emph{i.e.,} -0.2\% average accuracy.  When increasing the routing ratio to 50\% tokens, the inference throughput of $\gamma$-MoD-0.5 further improves  by up to +53.2\%.   Despite the significant efficiency gains,  the performance drop of  $\gamma$-MoD is still acceptable, \emph{i.e.,} -1.5\% average accuracy.  These results well validate the obvious benefits of $\gamma$-MoD in efficiency.

\subsubsection{Comparison with Existing MLLMs}

In Tab.~\ref{compare-sota}, we compare  MLLMs deployed by $\gamma$-MoD with both dense and sparse models on 9 benchmarks.  From it we can see $\gamma$-MoD can maintain the competitive performance on all benchmarks, while achieving  significant efficiency gains on LLaVA and LLaVA-HR.  Specifically, $\gamma$-MoD-LLaVA-HR (13B) can reach similar inference speed as LLaVA-HR (7B) while outperforming the latter on multiple benchmarks, \emph{e.g.,} +3.0\% on MMVet.  Compared to other dense MLLMs, similar merits of $\gamma$-MoD-LLaVA-HR can still be witnessed.  For example, $\gamma$-MoD-LLaVA-HR-7B not only obviously outperforms Qwen-VL    on GQA and VQAv2, but also demonstrates superior inference efficiency, \emph{i.e.,} 1.5 $\times$ speedup.   In addition, compared to existing sparse models, \emph{i.e.,} MoE-LLaVA~\citep{moellava}, our approaches also achieve better trade-off between performance and efficiency.  In particular, $\gamma$-MoD-LLaVA-HR (7B) outperforms MoE-LLaVA (5B) on 5 of 8 benchmarks, \emph{e.g.,} + 93 scores on MME, while still maintaining better efficiency, \emph{i.e.,} +28\% gains on inference speed. It is worth noting that although the parameter scale of MoE-LLaVA is smaller, its routing calculation often leads to higher latency.  More importantly,  MoE-LLaVA requires additional training stages for its MoE adaptation, which also consumes much more training data than our methods, \emph{i.e.,} 2.2M \textit{vs.} 1.1M.  Overall, these comparisons further confirm the effectiveness and efficiency of $\gamma$-MoD.


\subsubsection{Qualitative Analysis}
\begin{figure*}[t]
    \centering
    \includegraphics[width=1.\textwidth]{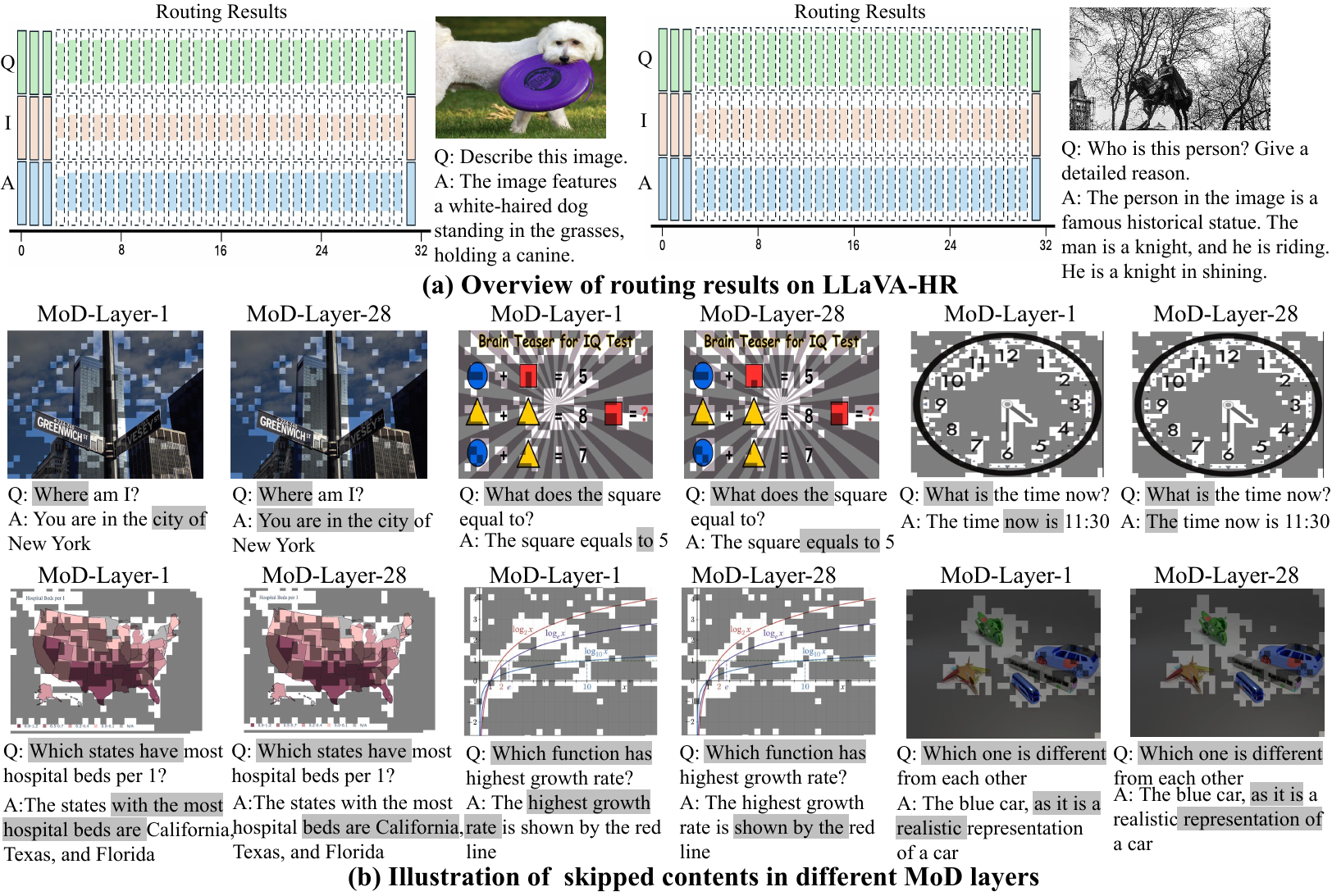}  
    \vspace{-2em}  
    \caption{\textbf{Visualization of routing results for different MoD layers.}   ``Q'', ``I'' and ``A'' denote the question, image and  response, respectively.  The skipped tokens in sub-figure (b) are colored in gray. }
    \label{fig4}
    \vspace{-1em}  
\end{figure*}
In Fig.~\ref{fig4},we visualize the routing ratio and the skipped content in both images and the corresponding conversations. The first observation from Fig.~\ref{fig4}.(a) is that question, image, and response tokens are routed in a consistent pattern: question tokens are mostly kept, while image tokens are the most redundant, and thus routed the most. In Fig.~\ref{fig4}.(b), we visualize the skipped content on images and texts. The gray portions of the images represent tokens that are skipped by the router, indicating that many regions in the images, such as background pixels, are redundant and do not provide critical information for understanding. Routing out these tokens allows the model to focus more on the white portions, which highlight the image regions or text parts that the model pays closer attention to. For example, in the middle of the first row with the IQ test example, the model can concentrate and spending more computations on the arithmetic and geometric aspects of the image, leading to a reasonable answer.
\section{Conclusion}
In this paper, we aim to overcome the efficiency problem in multimodal large language models (MLLMs) from the perspective of ``activated token''. 
In particular, we present $\gamma$-MoD, a novel mixture-of-depth adaptation strategy for  computationally efficient  MLLM.  In $\gamma$-MoD, an innovative metric is introduced to identify the redundant layers  for MoD deployment, namely \textit{rank of attention maps }(ARank). Moreover, $\gamma$-MoD also maximizes its benefit to MLLMs via two designs called \textit{shared vision-language router} and \textit{masked routing learning}. With these novel designs,  $\gamma$-MoD can obviously reduce computational costs of existing MLLMs while maintaining their performance. Extensive experiments on 9 multimodal benchmarks validate the efficiency and effectiveness. Besides, the great generalization ability of $\gamma$-MoD is also validated across different MLLMs.  
\paragraph{Acknowledgements:}
This work was supported by  the National Natural Science Foundation of China (No. 623B2088).

\bibliography{gamma_MoD}

\begin{thebibliography}{61}
\providecommand{\natexlab}[1]{#1}
\providecommand{\url}[1]{\texttt{#1}}
\expandafter\ifx\csname urlstyle\endcsname\relax
  \providecommand{\doi}[1]{doi: #1}\else
  \providecommand{\doi}{doi: \begingroup \urlstyle{rm}\Url}\fi

\bibitem[Abdin et~al.(2024)Abdin, Jacobs, Awan, Aneja, Awadallah, Awadalla, Bach, Bahree, Bakhtiari, Behl, et~al.]{phi3}
Marah Abdin, Sam~Ade Jacobs, Ammar~Ahmad Awan, Jyoti Aneja, Ahmed Awadallah, Hany Awadalla, Nguyen Bach, Amit Bahree, Arash Bakhtiari, Harkirat Behl, et~al.
\newblock Phi-3 technical report: A highly capable language model locally on your phone.
\newblock \emph{arXiv preprint arXiv:2404.14219}, 2024.

\bibitem[Achiam et~al.(2023)Achiam, Adler, Agarwal, Ahmad, Akkaya, Aleman, Almeida, Altenschmidt, Altman, Anadkat, et~al.]{GPT-4}
Josh Achiam, Steven Adler, Sandhini Agarwal, Lama Ahmad, Ilge Akkaya, Florencia~Leoni Aleman, Diogo Almeida, Janko Altenschmidt, Sam Altman, Shyamal Anadkat, et~al.
\newblock Gpt-4 technical report.
\newblock \emph{arXiv preprint arXiv:2303.08774}, 2023.

\bibitem[Adler et~al.(2024)Adler, Agarwal, Aithal, Anh, Bhattacharya, Brundyn, Casper, Catanzaro, Clay, Cohen, et~al.]{nemotron340B}
Bo~Adler, Niket Agarwal, Ashwath Aithal, Dong~H Anh, Pallab Bhattacharya, Annika Brundyn, Jared Casper, Bryan Catanzaro, Sharon Clay, Jonathan Cohen, et~al.
\newblock Nemotron-4 340b technical report.
\newblock \emph{arXiv preprint arXiv:2406.11704}, 2024.

\bibitem[Alayrac et~al.(2022)Alayrac, Donahue, Luc, Miech, Barr, Hasson, Lenc, Mensch, Millican, Reynolds, et~al.]{flamingo}
Jean-Baptiste Alayrac, Jeff Donahue, Pauline Luc, Antoine Miech, Iain Barr, Yana Hasson, Karel Lenc, Arthur Mensch, Katherine Millican, Malcolm Reynolds, et~al.
\newblock Flamingo: a visual language model for few-shot learning.
\newblock \emph{Advances in neural information processing systems}, 2022.

\bibitem[Almazrouei et~al.(2023)Almazrouei, Alobeidli, Alshamsi, Cappelli, Cojocaru, Debbah, Goffinet, Hesslow, Launay, Malartic, et~al.]{Falcon}
Ebtesam Almazrouei, Hamza Alobeidli, Abdulaziz Alshamsi, Alessandro Cappelli, Ruxandra Cojocaru, M{\'e}rouane Debbah, {\'E}tienne Goffinet, Daniel Hesslow, Julien Launay, Quentin Malartic, et~al.
\newblock The falcon series of open language models.
\newblock \emph{arXiv preprint arXiv:2311.16867}, 2023.

\bibitem[Bai et~al.(2023{\natexlab{a}})Bai, Bai, Chu, Cui, Dang, Deng, Fan, Ge, Han, Huang, et~al.]{Qwen}
Jinze Bai, Shuai Bai, Yunfei Chu, Zeyu Cui, Kai Dang, Xiaodong Deng, Yang Fan, Wenbin Ge, Yu~Han, Fei Huang, et~al.
\newblock Qwen technical report.
\newblock \emph{arXiv preprint arXiv:2309.16609}, 2023{\natexlab{a}}.

\bibitem[Bai et~al.(2023{\natexlab{b}})Bai, Bai, Yang, Wang, Tan, Wang, Lin, Zhou, and Zhou]{qwenvl}
Jinze Bai, Shuai Bai, Shusheng Yang, Shijie Wang, Sinan Tan, Peng Wang, Junyang Lin, Chang Zhou, and Jingren Zhou.
\newblock Qwen-vl: A frontier large vision-language model with versatile abilities.
\newblock \emph{arXiv preprint arXiv:2308.12966}, 2023{\natexlab{b}}.

\bibitem[Cai et~al.(2024{\natexlab{a}})Cai, Jiang, Wang, Tang, Kim, and Huang]{moesurvey}
Weilin Cai, Juyong Jiang, Fan Wang, Jing Tang, Sunghun Kim, and Jiayi Huang.
\newblock A survey on mixture of experts.
\newblock \emph{arXiv preprint arXiv:2407.06204}, 2024{\natexlab{a}}.

\bibitem[Cai et~al.(2024{\natexlab{b}})Cai, Cao, Chen, Chen, Chen, Chen, Chen, Chen, Chen, Chu, et~al.]{internLM}
Zheng Cai, Maosong Cao, Haojiong Chen, Kai Chen, Keyu Chen, Xin Chen, Xun Chen, Zehui Chen, Zhi Chen, Pei Chu, et~al.
\newblock Internlm2 technical report.
\newblock \emph{arXiv preprint arXiv:2403.17297}, 2024{\natexlab{b}}.

\bibitem[Chen et~al.(2023)Chen, Zhu, Shen, Li, Liu, Zhang, Krishnamoorthi, Chandra, Xiong, and Elhoseiny]{minigpt}
Jun Chen, Deyao Zhu, Xiaoqian Shen, Xiang Li, Zechun Liu, Pengchuan Zhang, Raghuraman Krishnamoorthi, Vikas Chandra, Yunyang Xiong, and Mohamed Elhoseiny.
\newblock Minigpt-v2: large language model as a unified interface for vision-language multi-task learning.
\newblock \emph{arXiv preprint arXiv:2310.09478}, 2023.

\bibitem[Chen et~al.(2024{\natexlab{a}})Chen, Wang, Tian, Ye, Gao, Cui, Tong, Hu, Luo, Ma, Ma, Wang, Dong, Yan, Guo, He, Shi, Jin, Xu, Wang, Wei, Li, Zhang, Zhang, Cai, Wen, Yan, Dou, Lu, Zhu, Lu, Lin, Qiao, Dai, and Wang]{internvl1.5}
Zhe Chen, Weiyun Wang, Hao Tian, Shenglong Ye, Zhangwei Gao, Erfei Cui, Wenwen Tong, Kongzhi Hu, Jiapeng Luo, Zheng Ma, Ji~Ma, Jiaqi Wang, Xiaoyi Dong, Hang Yan, Hewei Guo, Conghui He, Botian Shi, Zhenjiang Jin, Chao Xu, Bin Wang, Xingjian Wei, Wei Li, Wenjian Zhang, Bo~Zhang, Pinlong Cai, Licheng Wen, Xiangchao Yan, Min Dou, Lewei Lu, Xizhou Zhu, Tong Lu, Dahua Lin, Yu~Qiao, Jifeng Dai, and Wenhai Wang.
\newblock How far are we to gpt-4v? closing the gap to commercial multimodal models with open-source suites, 2024{\natexlab{a}}.

\bibitem[Chen et~al.(2024{\natexlab{b}})Chen, Wu, Wang, Su, Chen, Xing, Zhong, Zhang, Zhu, Lu, et~al.]{internVL}
Zhe Chen, Jiannan Wu, Wenhai Wang, Weijie Su, Guo Chen, Sen Xing, Muyan Zhong, Qinglong Zhang, Xizhou Zhu, Lewei Lu, et~al.
\newblock Internvl: Scaling up vision foundation models and aligning for generic visual-linguistic tasks.
\newblock In \emph{Proceedings of the IEEE/CVF Conference on Computer Vision and Pattern Recognition}, pp.\  24185--24198, 2024{\natexlab{b}}.

\bibitem[Dai et~al.(2024)Dai, Deng, Zhao, Xu, Gao, Chen, Li, Zeng, Yu, Wu, et~al.]{deepseekmoe}
Damai Dai, Chengqi Deng, Chenggang Zhao, RX~Xu, Huazuo Gao, Deli Chen, Jiashi Li, Wangding Zeng, Xingkai Yu, Y~Wu, et~al.
\newblock Deepseekmoe: Towards ultimate expert specialization in mixture-of-experts language models.
\newblock \emph{arXiv preprint arXiv:2401.06066}, 2024.

\bibitem[Dai et~al.(2023)Dai, Li, Li, Tiong, Zhao, Wang, Li, Fung, and Hoi]{InstructBLIP}
Wenliang Dai, Junnan Li, Dongxu Li, Anthony Meng~Huat Tiong, Junqi Zhao, Weisheng Wang, Boyang Li, Pascale Fung, and Steven Hoi.
\newblock Instructblip: Towards general-purpose vision-language models with instruction tuning, 2023.

\bibitem[Dong et~al.(2023)Dong, Han, Peng, Qi, Ge, Yang, Zhao, Sun, Zhou, Wei, et~al.]{dreamllm}
Runpei Dong, Chunrui Han, Yuang Peng, Zekun Qi, Zheng Ge, Jinrong Yang, Liang Zhao, Jianjian Sun, Hongyu Zhou, Haoran Wei, et~al.
\newblock Dreamllm: Synergistic multimodal comprehension and creation.
\newblock \emph{arXiv preprint arXiv:2309.11499}, 2023.

\bibitem[Dubey et~al.(2024)Dubey, Jauhri, Pandey, Kadian, Al-Dahle, Letman, Mathur, Schelten, Yang, Fan, et~al.]{llama405B}
Abhimanyu Dubey, Abhinav Jauhri, Abhinav Pandey, Abhishek Kadian, Ahmad Al-Dahle, Aiesha Letman, Akhil Mathur, Alan Schelten, Amy Yang, Angela Fan, et~al.
\newblock The llama 3 herd of models.
\newblock \emph{arXiv preprint arXiv:2407.21783}, 2024.

\bibitem[Fu et~al.(2024)Fu, Chen, Shen, Qin, Zhang, Lin, Yang, Zheng, Li, Sun, Wu, and Ji]{MME}
Chaoyou Fu, Peixian Chen, Yunhang Shen, Yulei Qin, Mengdan Zhang, Xu~Lin, Jinrui Yang, Xiawu Zheng, Ke~Li, Xing Sun, Yunsheng Wu, and Rongrong Ji.
\newblock Mme: A comprehensive evaluation benchmark for multimodal large language models, 2024.

\bibitem[Fyodorov et~al.(2000)Fyodorov, Winter, and Francez]{logicinfer}
Yaroslav Fyodorov, Yoad Winter, and Nissim Francez.
\newblock A natural logic inference system.
\newblock In \emph{Proceedings of the 2nd workshop on inference in computational semantics (ICoS-2)}, 2000.

\bibitem[Ge et~al.(2023)Ge, Ge, Zeng, Wang, and Shan]{seed}
Yuying Ge, Yixiao Ge, Ziyun Zeng, Xintao Wang, and Ying Shan.
\newblock Planting a seed of vision in large language model, 2023.

\bibitem[G.H.Goulb \& C.Reinsch(1971)G.H.Goulb and C.Reinsch]{svd}
G.H.Goulb and C.Reinsch.
\newblock Singular value decomposition and least squares solutions.
\newblock In \emph{Handbook for Automatic Computation: Volume II: Linear Algebra}, pp.\  134--151. Springer, 1971.

\bibitem[Goyal et~al.(2017)Goyal, Khot, Summers-Stay, Batra, and Parikh]{vqav2}
Yash Goyal, Tejas Khot, Douglas Summers-Stay, Dhruv Batra, and Devi Parikh.
\newblock Making the v in vqa matter: Elevating the role of image understanding in visual question answering, 2017.

\bibitem[Gurari et~al.(2018)Gurari, Li, Stangl, Guo, Lin, Grauman, Luo, and Bigham]{vizwiz}
Danna Gurari, Qing Li, Abigale~J. Stangl, Anhong Guo, Chi Lin, Kristen Grauman, Jiebo Luo, and Jeffrey~P. Bigham.
\newblock Vizwiz grand challenge: Answering visual questions from blind people, 2018.

\bibitem[Hudson \& Manning(2019)Hudson and Manning]{gqa}
Drew~A. Hudson and Christopher~D. Manning.
\newblock Gqa: A new dataset for real-world visual reasoning and compositional question answering, 2019.

\bibitem[Jiang et~al.(2024)Jiang, Sablayrolles, Roux, Mensch, Savary, Bamford, Chaplot, Casas, Hanna, Bressand, et~al.]{mixtral}
Albert~Q Jiang, Alexandre Sablayrolles, Antoine Roux, Arthur Mensch, Blanche Savary, Chris Bamford, Devendra~Singh Chaplot, Diego de~las Casas, Emma~Bou Hanna, Florian Bressand, et~al.
\newblock Mixtral of experts.
\newblock \emph{arXiv preprint arXiv:2401.04088}, 2024.

\bibitem[Jin et~al.(2024)Jin, Li, Liu, Gu, Wu, Jiang, He, Zhao, Tan, Gan, et~al.]{efficient_mllm}
Yizhang Jin, Jian Li, Yexin Liu, Tianjun Gu, Kai Wu, Zhengkai Jiang, Muyang He, Bo~Zhao, Xin Tan, Zhenye Gan, et~al.
\newblock Efficient multimodal large language models: A survey.
\newblock \emph{arXiv preprint arXiv:2405.10739}, 2024.

\bibitem[Kim et~al.(2024)Kim, Gao, Hsu, Shen, and Jin]{tokenfusion}
Minchul Kim, Shangqian Gao, Yen-Chang Hsu, Yilin Shen, and Hongxia Jin.
\newblock Token fusion: Bridging the gap between token pruning and token merging.
\newblock In \emph{Proceedings of the IEEE/CVF Winter Conference on Applications of Computer Vision}, pp.\  1383--1392, 2024.

\bibitem[Lauren{\c{c}}on et~al.(2024)Lauren{\c{c}}on, Saulnier, Tronchon, Bekman, Singh, Lozhkov, Wang, Karamcheti, Rush, Kiela, et~al.]{i80B}
Hugo Lauren{\c{c}}on, Lucile Saulnier, L{\'e}o Tronchon, Stas Bekman, Amanpreet Singh, Anton Lozhkov, Thomas Wang, Siddharth Karamcheti, Alexander Rush, Douwe Kiela, et~al.
\newblock Obelics: An open web-scale filtered dataset of interleaved image-text documents.
\newblock \emph{Advances in Neural Information Processing Systems}, 36, 2024.

\bibitem[Li et~al.(2024{\natexlab{a}})Li, Zhang, Guo, Zhang, Li, Zhang, Zhang, Li, Liu, and Li]{LLaVA-Onevision}
Bo~Li, Yuanhan Zhang, Dong Guo, Renrui Zhang, Feng Li, Hao Zhang, Kaichen Zhang, Yanwei Li, Ziwei Liu, and Chunyuan Li.
\newblock Llava-onevision: Easy visual task transfer.
\newblock \emph{arXiv preprint arXiv:2408.03326}, 2024{\natexlab{a}}.

\bibitem[Li et~al.(2024{\natexlab{b}})Li, Zhang, Wang, Zhong, Chen, Chu, Liu, and Jia]{miniGemini}
Yanwei Li, Yuechen Zhang, Chengyao Wang, Zhisheng Zhong, Yixin Chen, Ruihang Chu, Shaoteng Liu, and Jiaya Jia.
\newblock Mini-gemini: Mining the potential of multi-modality vision language models.
\newblock \emph{arXiv preprint arXiv:2403.18814}, 2024{\natexlab{b}}.

\bibitem[Li et~al.(2023)Li, Du, Zhou, Wang, Zhao, and Wen]{pope}
Yifan Li, Yifan Du, Kun Zhou, Jinpeng Wang, Wayne~Xin Zhao, and Ji-Rong Wen.
\newblock Evaluating object hallucination in large vision-language models, 2023.

\bibitem[Lin et~al.(2024{\natexlab{a}})Lin, Tang, Ye, Cui, Zhu, Jin, Huang, Zhang, Pang, Ning, and Yuan]{moellava}
Bin Lin, Zhenyu Tang, Yang Ye, Jiaxi Cui, Bin Zhu, Peng Jin, Jinfa Huang, Junwu Zhang, Yatian Pang, Munan Ning, and Li~Yuan.
\newblock Moe-llava: Mixture of experts for large vision-language models, 2024{\natexlab{a}}.

\bibitem[Lin et~al.(2024{\natexlab{b}})Lin, Yin, Ping, Molchanov, Shoeybi, and Han]{lin2024vila}
Ji~Lin, Hongxu Yin, Wei Ping, Pavlo Molchanov, Mohammad Shoeybi, and Song Han.
\newblock Vila: On pre-training for visual language models.
\newblock In \emph{Proceedings of the IEEE/CVF Conference on Computer Vision and Pattern Recognition}, pp.\  26689--26699, 2024{\natexlab{b}}.

\bibitem[Lin et~al.(2020)Lin, Ji, Wang, Zhang, Zhang, Tian, and Shao]{HRank}
Mingbao Lin, Rongrong Ji, Yan Wang, Yichen Zhang, Baochang Zhang, Yonghong Tian, and Ling Shao.
\newblock Hrank: Filter pruning using high-rank feature map.
\newblock In \emph{Proceedings of the IEEE/CVF conference on computer vision and pattern recognition}, pp.\  1529--1538, 2020.

\bibitem[Liu et~al.(2024{\natexlab{a}})Liu, Feng, Wang, Wang, Liu, Zhao, Dengr, Ruan, Dai, Guo, et~al.]{deepseekmoev2}
Aixin Liu, Bei Feng, Bin Wang, Bingxuan Wang, Bo~Liu, Chenggang Zhao, Chengqi Dengr, Chong Ruan, Damai Dai, Daya Guo, et~al.
\newblock Deepseek-v2: A strong, economical, and efficient mixture-of-experts language model.
\newblock \emph{arXiv preprint arXiv:2405.04434}, 2024{\natexlab{a}}.

\bibitem[Liu et~al.(2024{\natexlab{b}})Liu, Li, Li, and Lee]{llava1.5}
Haotian Liu, Chunyuan Li, Yuheng Li, and Yong~Jae Lee.
\newblock Improved baselines with visual instruction tuning.
\newblock In \emph{Proceedings of the IEEE/CVF Conference on Computer Vision and Pattern Recognition}, pp.\  26296--26306, 2024{\natexlab{b}}.

\bibitem[Liu et~al.(2024{\natexlab{c}})Liu, Li, Li, Li, Zhang, Shen, and Lee]{llava-next}
Haotian Liu, Chunyuan Li, Yuheng Li, Bo~Li, Yuanhan Zhang, Sheng Shen, and Yong~Jae Lee.
\newblock Llava-next: Improved reasoning, ocr, and world knowledge, January 2024{\natexlab{c}}.

\bibitem[Liu et~al.(2024{\natexlab{d}})Liu, Li, Wu, and Lee]{LLaVA}
Haotian Liu, Chunyuan Li, Qingyang Wu, and Yong~Jae Lee.
\newblock Visual instruction tuning.
\newblock \emph{Advances in neural information processing systems}, 36, 2024{\natexlab{d}}.

\bibitem[Liu et~al.(2024{\natexlab{e}})Liu, Duan, Zhang, Li, Zhang, Zhao, Yuan, Wang, He, Liu, Chen, and Lin]{MMB}
Yuan Liu, Haodong Duan, Yuanhan Zhang, Bo~Li, Songyang Zhang, Wangbo Zhao, Yike Yuan, Jiaqi Wang, Conghui He, Ziwei Liu, Kai Chen, and Dahua Lin.
\newblock Mmbench: Is your multi-modal model an all-around player?, 2024{\natexlab{e}}.

\bibitem[Liu et~al.(2024{\natexlab{f}})Liu, Tan, Wang, Neiswanger, Tao, Li, Koto, Wang, Sun, Pangarkar, Fan, Gu, Miller, Ma, Tang, Ranjan, Zhuang, He, Wang, Deng, Algayres, Li, Shen, Nakov, and Xing]{K265B}
Zhengzhong Liu, Bowen Tan, Hongyi Wang, Willie Neiswanger, Tianhua Tao, Haonan Li, Fajri Koto, Yuqi Wang, Suqi Sun, Omkar Pangarkar, Richard Fan, Yi~Gu, Victor Miller, Liqun Ma, Liping Tang, Nikhil Ranjan, Yonghao Zhuang, Guowei He, Renxi Wang, Mingkai Deng, Robin Algayres, Yuanzhi Li, Zhiqiang Shen, Preslav Nakov, and Eric Xing.
\newblock Llm360 k2-65b: Scaling up fully transparent open-source llms.
\newblock 2024{\natexlab{f}}.

\bibitem[Lu et~al.(2022)Lu, Mishra, Xia, Qiu, Chang, Zhu, Tafjord, Clark, and Kalyan]{sqa}
Pan Lu, Swaroop Mishra, Tony Xia, Liang Qiu, Kai-Wei Chang, Song-Chun Zhu, Oyvind Tafjord, Peter Clark, and Ashwin Kalyan.
\newblock Learn to explain: Multimodal reasoning via thought chains for science question answering, 2022.

\bibitem[Luo et~al.(2024{\natexlab{a}})Luo, Yang, Dou, Wang, Dai, Qiao, and Zhu]{monointernvl}
Gen Luo, Xue Yang, Wenhan Dou, Zhaokai Wang, Jifeng Dai, Yu~Qiao, and Xizhou Zhu.
\newblock Mono-internvl: Pushing the boundaries of monolithic multimodal large language models with endogenous visual pre-training.
\newblock \emph{arXiv preprint arXiv:2410.08202}, 2024{\natexlab{a}}.

\bibitem[Luo et~al.(2024{\natexlab{b}})Luo, Zhou, Zhang, Zheng, Sun, and Ji]{llava-hr}
Gen Luo, Yiyi Zhou, Yuxin Zhang, Xiawu Zheng, Xiaoshuai Sun, and Rongrong Ji.
\newblock Feast your eyes: Mixture-of-resolution adaptation for multimodal large language models.
\newblock \emph{arXiv preprint arXiv:2403.03003}, 2024{\natexlab{b}}.

\bibitem[McKinzie et~al.(2024)McKinzie, Gan, Fauconnier, Dodge, Zhang, Dufter, Shah, Du, Peng, Weers, et~al.]{mm1}
Brandon McKinzie, Zhe Gan, Jean-Philippe Fauconnier, Sam Dodge, Bowen Zhang, Philipp Dufter, Dhruti Shah, Xianzhi Du, Futang Peng, Floris Weers, et~al.
\newblock Mm1: Methods, analysis \& insights from multimodal llm pre-training.
\newblock \emph{arXiv preprint arXiv:2403.09611}, 2024.

\bibitem[Pal et~al.(2024)Pal, Karkhanis, Dooley, Roberts, Naidu, and White]{smaug72B}
Arka Pal, Deep Karkhanis, Samuel Dooley, Manley Roberts, Siddartha Naidu, and Colin White.
\newblock Smaug: Fixing failure modes of preference optimisation with dpo-positive.
\newblock \emph{arXiv preprint arXiv:2402.13228}, 2024.

\bibitem[Paperno et~al.(2016)Paperno, Kruszewski, Lazaridou, Pham, Bernardi, Pezzelle, Baroni, Boleda, and Fern{\'a}ndez]{lambada}
Denis Paperno, Germ{\'a}n Kruszewski, Angeliki Lazaridou, Quan~Ngoc Pham, Raffaella Bernardi, Sandro Pezzelle, Marco Baroni, Gemma Boleda, and Raquel Fern{\'a}ndez.
\newblock The lambada dataset: Word prediction requiring a broad discourse context.
\newblock \emph{arXiv preprint arXiv:1606.06031}, 2016.

\bibitem[Raposo et~al.(2024)Raposo, Ritter, Richards, Lillicrap, Humphreys, and Santoro]{mod}
David Raposo, Sam Ritter, Blake Richards, Timothy Lillicrap, Peter~Conway Humphreys, and Adam Santoro.
\newblock Mixture-of-depths: Dynamically allocating compute in transformer-based language models.
\newblock \emph{arXiv preprint arXiv:2404.02258}, 2024.

\bibitem[Rasheed et~al.(2024)Rasheed, Maaz, Khan, and Khan]{llava-pp}
Hanoona Rasheed, Muhammad Maaz, Salman Khan, and Fahad~S. Khan.
\newblock Llava++: Extending visual capabilities with llama-3 and phi-3, 2024.

\bibitem[Reddy et~al.(2019)Reddy, Chen, and Manning]{readingcomphrehension}
Siva Reddy, Danqi Chen, and Christopher~D Manning.
\newblock Coqa: A conversational question answering challenge.
\newblock \emph{Transactions of the Association for Computational Linguistics}, 7:\penalty0 249--266, 2019.

\bibitem[Shen et~al.(2023)Shen, Tao, Ma, Neiswanger, Hestness, Vassilieva, Soboleva, and Xing]{SlimPajama-DC}
Zhiqiang Shen, Tianhua Tao, Liqun Ma, Willie Neiswanger, Joel Hestness, Natalia Vassilieva, Daria Soboleva, and Eric Xing.
\newblock Slimpajama-dc: Understanding data combinations for llm training.
\newblock \emph{arXiv preprint arXiv:2309.10818}, 2023.

\bibitem[Singh et~al.(2019)Singh, Natarajan, Shah, Jiang, Chen, Batra, Parikh, and Rohrbach]{textvqa}
Amanpreet Singh, Vivek Natarajan, Meet Shah, Yu~Jiang, Xinlei Chen, Dhruv Batra, Devi Parikh, and Marcus Rohrbach.
\newblock Towards vqa models that can read.
\newblock In \emph{Proceedings of the IEEE/CVF conference on computer vision and pattern recognition}, pp.\  8317--8326, 2019.

\bibitem[Sun et~al.(2024)Sun, Cui, Zhang, Zhang, Yu, Wang, Rao, Liu, Huang, and Wang]{emu2}
Quan Sun, Yufeng Cui, Xiaosong Zhang, Fan Zhang, Qiying Yu, Yueze Wang, Yongming Rao, Jingjing Liu, Tiejun Huang, and Xinlong Wang.
\newblock Generative multimodal models are in-context learners.
\newblock In \emph{Proceedings of the IEEE/CVF Conference on Computer Vision and Pattern Recognition}, pp.\  14398--14409, 2024.

\bibitem[Tong et~al.(2024)Tong, Brown, Wu, Woo, Middepogu, Akula, Yang, Yang, Iyer, Pan, et~al.]{Cambrian-1}
Shengbang Tong, Ellis Brown, Penghao Wu, Sanghyun Woo, Manoj Middepogu, Sai~Charitha Akula, Jihan Yang, Shusheng Yang, Adithya Iyer, Xichen Pan, et~al.
\newblock Cambrian-1: A fully open, vision-centric exploration of multimodal llms.
\newblock \emph{arXiv preprint arXiv:2406.16860}, 2024.

\bibitem[Touvron et~al.(2023)Touvron, Lavril, Izacard, Martinet, Lachaux, Lacroix, Rozi{\`e}re, Goyal, Hambro, Azhar, et~al.]{LLaMA}
Hugo Touvron, Thibaut Lavril, Gautier Izacard, Xavier Martinet, Marie-Anne Lachaux, Timoth{\'e}e Lacroix, Baptiste Rozi{\`e}re, Naman Goyal, Eric Hambro, Faisal Azhar, et~al.
\newblock Llama: Open and efficient foundation language models.
\newblock \emph{arXiv preprint arXiv:2302.13971}, 2023.

\bibitem[Xie et~al.(2024)Xie, Mao, Bai, Zhang, Wang, Lin, Gu, Chen, Yang, and Shou]{showo}
Jinheng Xie, Weijia Mao, Zechen Bai, David~Junhao Zhang, Weihao Wang, Kevin~Qinghong Lin, Yuchao Gu, Zhijie Chen, Zhenheng Yang, and Mike~Zheng Shou.
\newblock Show-o: One single transformer to unify multimodal understanding and generation.
\newblock \emph{arXiv preprint arXiv:2408.12528}, 2024.

\bibitem[Xue et~al.(2024)Xue, Zheng, Fu, Ni, Zheng, Zhou, and You]{openmoe}
Fuzhao Xue, Zian Zheng, Yao Fu, Jinjie Ni, Zangwei Zheng, Wangchunshu Zhou, and Yang You.
\newblock Openmoe: An early effort on open mixture-of-experts language models.
\newblock \emph{arXiv preprint arXiv:2402.01739}, 2024.

\bibitem[Yang et~al.(2024)Yang, Yang, Hui, Zheng, Yu, Zhou, Li, Li, Liu, Huang, et~al.]{Qwen72B}
An~Yang, Baosong Yang, Binyuan Hui, Bo~Zheng, Bowen Yu, Chang Zhou, Chengpeng Li, Chengyuan Li, Dayiheng Liu, Fei Huang, et~al.
\newblock Qwen2 technical report.
\newblock \emph{arXiv preprint arXiv:2407.10671}, 2024.

\bibitem[Ye et~al.(2023)Ye, Xu, Xu, Ye, Yan, Zhou, Wang, Hu, Shi, Shi, et~al.]{mPLUG}
Qinghao Ye, Haiyang Xu, Guohai Xu, Jiabo Ye, Ming Yan, Yiyang Zhou, Junyang Wang, Anwen Hu, Pengcheng Shi, Yaya Shi, et~al.
\newblock mplug-owl: Modularization empowers large language models with multimodality.
\newblock \emph{arXiv preprint arXiv:2304.14178}, 2023.

\bibitem[Yu et~al.(2023)Yu, Yang, Li, Wang, Lin, Liu, Wang, and Wang]{mmvet}
Weihao Yu, Zhengyuan Yang, Linjie Li, Jianfeng Wang, Kevin Lin, Zicheng Liu, Xinchao Wang, and Lijuan Wang.
\newblock Mm-vet: Evaluating large multimodal models for integrated capabilities, 2023.

\bibitem[Yue et~al.(2024)Yue, Ni, Zhang, Zheng, Liu, Zhang, Stevens, Jiang, Ren, Sun, Wei, Yu, Yuan, Sun, Yin, Zheng, Yang, Liu, Huang, Sun, Su, and Chen]{MMMU}
Xiang Yue, Yuansheng Ni, Kai Zhang, Tianyu Zheng, Ruoqi Liu, Ge~Zhang, Samuel Stevens, Dongfu Jiang, Weiming Ren, Yuxuan Sun, Cong Wei, Botao Yu, Ruibin Yuan, Renliang Sun, Ming Yin, Boyuan Zheng, Zhenzhu Yang, Yibo Liu, Wenhao Huang, Huan Sun, Yu~Su, and Wenhu Chen.
\newblock Mmmu: A massive multi-discipline multimodal understanding and reasoning benchmark for expert agi, 2024.

\bibitem[Zhou et~al.(2024)Zhou, Yu, Babu, Tirumala, Yasunaga, Shamis, Kahn, Ma, Zettlemoyer, and Levy]{transfusion}
Chunting Zhou, Lili Yu, Arun Babu, Kushal Tirumala, Michihiro Yasunaga, Leonid Shamis, Jacob Kahn, Xuezhe Ma, Luke Zettlemoyer, and Omer Levy.
\newblock Transfusion: Predict the next token and diffuse images with one multi-modal model.
\newblock \emph{arXiv preprint arXiv:2408.11039}, 2024.

\bibitem[Ziegler et~al.(2019)Ziegler, Stiennon, Wu, Brown, Radford, Amodei, Christiano, and Irving]{helloswag}
Daniel~M Ziegler, Nisan Stiennon, Jeffrey Wu, Tom~B Brown, Alec Radford, Dario Amodei, Paul Christiano, and Geoffrey Irving.
\newblock Fine-tuning language models from human preferences.
\newblock \emph{arXiv preprint arXiv:1909.08593}, 2019.

\end{thebibliography}
\bibliographystyle{arxiv_bib_style}


\end{document}